\documentclass[10pt,twocolumn,letterpaper]{article}

\usepackage{iccv}              

\usepackage{multirow}
\usepackage{multicol}
\usepackage{makecell}
\usepackage{graphicx}
\usepackage{booktabs}
\usepackage{colortbl}
\usepackage{amsmath}
\usepackage{amssymb}
\usepackage{bbding}
\usepackage[accsupp]{axessibility}  
\usepackage{tablefootnote}

\definecolor{iccvblue}{rgb}{0.21,0.49,0.74}
\usepackage[pagebackref,breaklinks,colorlinks,allcolors=iccvblue]{hyperref}

\def\paperID{2244} 
\def\confName{ICCV}
\def\confYear{2025}

\title{Hierarchical Event Memory for Accurate and Low-latency Online Video Temporal Grounding}

\author{Minghang Zheng$^1$ \quad Yuxin Peng$^1$ \quad Benyuan Sun$^3$ \quad Yi Yang$^3$ \quad Yang Liu$^{1,2}$\thanks{Corresponding author}\\
$^1$Wangxuan Institute of Computer Technology, Peking University\\
$^2$State Key Laboratory of General Artificial Intelligence, Peking University\\
$^3$Central Media Technology Institute, Huawei\\
{\tt\small \{minghang,pengyuxin,yangliu\}@pku.edu.cn} \\ 
{\tt\small \{sunbenyuan,yangyi16\}@huawei.com}
}

\begin{document}
\maketitle
\begin{abstract}
In this paper, we tackle the task of online video temporal grounding (OnVTG), which requires the model to locate events related to a given text query within a video stream. Unlike regular video temporal grounding, OnVTG requires the model to make predictions without observing future frames. As online videos are streaming inputs and can go on indefinitely, it is impractical and inefficient to store all historical inputs. The existing OnVTG models employ memory to store recent historical video frame features and predict scores indicating whether the current frame corresponds to the start or end time of the target event. However, these methods lack effective event modeling and cannot retain long-term historical information, leading to low performance. To tackle these challenges, we propose a hierarchical event memory for OnVTG. We propose an event-based OnVTG framework that makes predictions based on event proposals that model event-level information with various durations. To preserve historically valuable event information, we introduce a hierarchical event memory that retains historical events, allowing the model to access both recent and long-term information. To enable the real-time prediction, we further propose a future prediction branch that predicts whether the target event will occur shortly and further regresses the start time of the event. We achieve state-of-the-art performance on the TACoS, ActivityNet Captions, and MAD datasets. Code is available at \url{https://github.com/minghangz/OnVTG}.
\end{abstract}
\section{Introduction}
\label{sec:intro}

\begin{figure}[t]
    \centering
    \includegraphics[width=\linewidth]{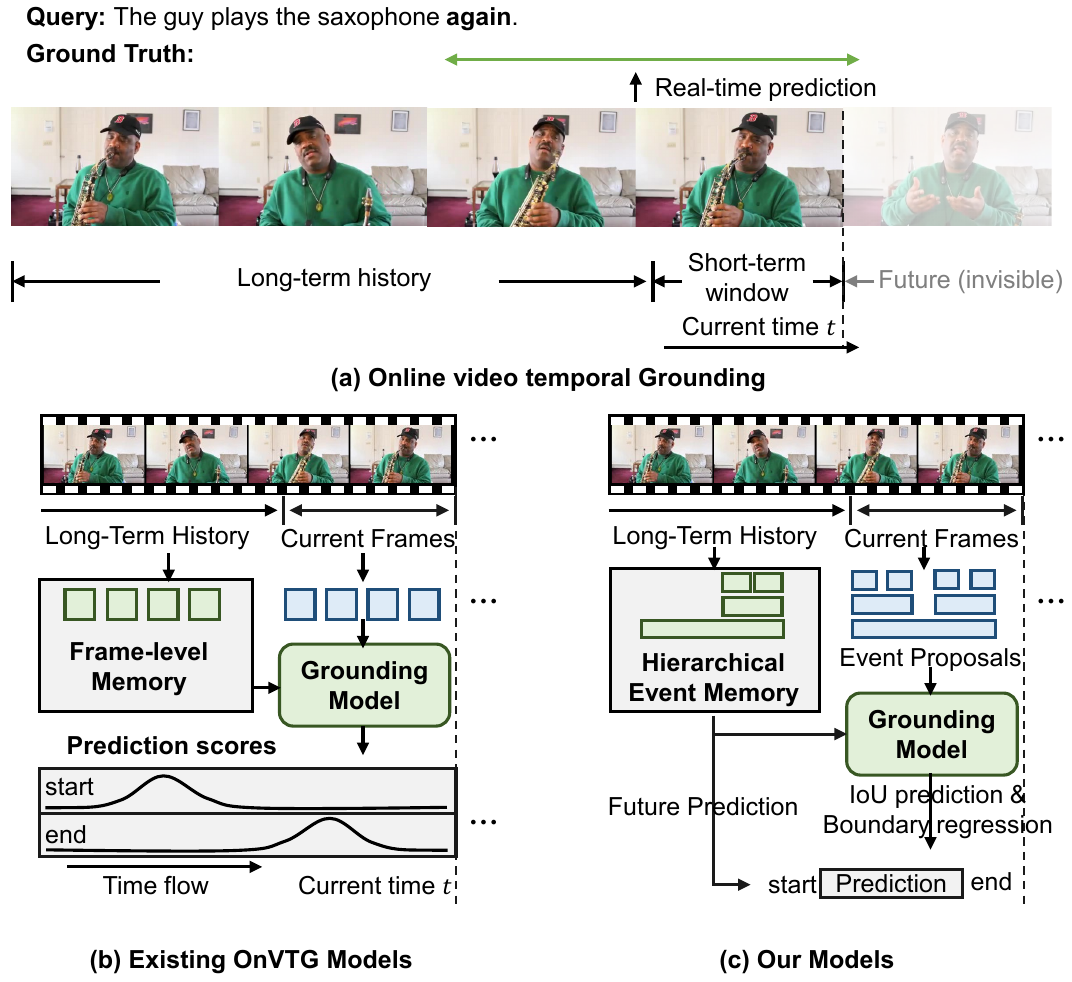}
    \caption{(a) The online video temporal grounding. (b) Existing work employs memory to store recent historical video frames and predicts scores indicating whether the current frame corresponds to the start or end time of the target event. (c) We propose a hierarchical event memory that retains historical events at different scales, allowing the model to acquire long-term historical information and make predictions based on event proposals.}
    \label{fig:teaser}
\end{figure}

In recent decades, the number of online videos, such as surveillance and live streaming, has increased significantly. As shown in Figure~\ref{fig:teaser} (a), online video temporal grounding (OnVTG)~\cite{gan2023temporal} can localize the location of events described by natural language queries within online videos in real time, showing potential applications in areas such as alert in surveillance~\cite{surveillance,E220173} and online cross-media retrieval~\cite{Peng2017AnOO,E210455,E210093}.

Unlike the traditional video temporal grounding~\cite{gao2017tall,zhang2020learning,wang2022negative,li2024momentdiff,mun2020local,liu2024towards,xiao2024bridging,wang2024bridge,Croitoru_2023_ICCV} method, the OnVTG model can only observe the video from $0$ to $t$ at the current timestamp $t$, and it should make predictions promptly when the target event occurs. As online videos are streaming inputs and can go on indefinitely, it is impractical and inefficient to store all historical inputs. Thus, most existing online video understanding methods~\cite{gan2023temporal,reza2025hat,kim2022sliding,ting2023hoi} only retain the latest historical information using a fixed-size memory. As shown in Figure~\ref{fig:teaser}(b), the existing OnVTG model~\cite{gan2023temporal} employs memory to store recent historical video frame features and predicts scores indicating whether the current frame corresponds to the start or end time of the target event. However, these methods have the following limitations.

\textit{First, video temporal grounding requires identifying target events of various durations within the video.} 
For example, for a person pushing open a door, the duration of the target event can be very short, but for a person playing the saxophone, the duration of the target event can be long. However, such per-frame score prediction methods lack effective event modeling, leading to poor performance. \textit{Secondly, the target event may require long-term valuable historical information for accurate localization.}  For example, in Figure~\ref{fig:teaser}(a), to locate the query `the guy plays the saxophone again', the model needs the information about when he first plays the saxophone, which happens at the beginning of the video. Previous methods typically employed a first-in-first-out principle to update the memory, resulting in the memory being filled with redundant content and valuable event information being removed from the memory when long-duration similar frames appear.


To address the aforementioned issues, we propose a hierarchical event memory for online video temporal grounding, as illustrated in Figure~\ref{fig:teaser}(c). \textit{First, we propose an event-based OnVTG framework to model complete event-level information with various durations.} We construct event proposals with various durations using a segment tree structure, classify whether these proposals match the query, and further regress the temporal boundaries of the matched proposals. \textit{Secondly, we propose a hierarchical event memory that retains long-term historical events.} The ideal memory should (1) retain long-term historical event information and (2) preserve valuable low-redundant information within a limited size. To achieve this, (1) we create a hierarchical memory where small-scale memory stores recent information, while large-scale memory stores long-term information. (2) We propose a dynamic memory size configuration for different scales, ensuring that those scales with more positive events have larger memory sizes. We also remove redundant events in the memory by proposing an adaptive memory update rule.


However, solely relying on event proposals for predictions can lead to delays in predicting the start of events since the model can only obtain a complete event proposal when the event is about to end. To solve this problem, we further propose a future prediction branch that predicts whether the target event will occur in the near future and further regresses the start time of the event. This enables our model to predict in two ways: one is to predict the start and end times based on the event proposal, and the other is to predict the start time using the future prediction branch and the end time using the event proposal. The former achieves higher performance by observing the entire event before making predictions, while the latter can make predictions when the event begins, achieving less start time prediction latency. Through hierarchical event memory and future prediction, we achieve significantly better performance and lower latency than existing methods, attaining state-of-the-art performance on the TACoS~\cite{rohrbach2014coherent}, ActivityNet~\cite{krishna2017dense}, and MAD~\cite{soldan2022mad} datasets.

Our contributions are summarized as follows: (1) We propose a novel pipeline for online video temporal grounding based on hierarchical event memory, enabling a more accurate and low-latency prediction. (2) We propose the dynamic memory size configuration and adaptive memory updating to enable the memory to preserve more valuable information within a limited size and reduce redundant information in the memory. (3) We propose a future prediction branch to address the issue of delays in model predictions. (4) We achieve state-of-the-art performance on the MAD, ActivityNet, and TACoS datasets.

\section{Related Work}
\label{sec:relatedwork}

\subsection{Offline Video Temporal Grounding}

Video temporal grounding (VTG) aims to localize the most relevant segments in untrimmed videos based on a given natural language query. Most of the existing work focuses on offline videos, where the model makes predictions after all the videos are inputted. These approaches can be categorized into two types: proposal-free~\cite{mun2020local,liu2024towards,xiao2024bridging,SPL_2023_ACL,VMR_2023_CVPR,huang2022emb} and proposal-based~\cite{gao2017tall,zhang2020learning,wang2022negative,li2024momentdiff,zheng-etal-2024-training,Luo_2024_WACV,TRM_2023_AAAI,CPL_2022_CVPR,CNM_2022_AAAI} methods. 
Proposal-free methods directly predict the target moment's start and end temporal boundaries. 
Proposal-based methods demonstrate better performance by modeling event proposals of varying lengths and positions within the video. The proposal-based method explicitly models event proposals in the video and typically achieves better performance. 
However, these proposal-based methods cannot be directly applied to online videos. Firstly, the number of these proposals scales quadratically with the video stream input, which is redundant and inefficient. Besides, the delays in predicting the start of events can be large since the model can only obtain a complete event proposal when the event is about to end.

\subsection{Online Video Understanding}
Existing research on online videos mainly includes online video temporal grounding~\cite{gan2023temporal}, action detection~\cite{de2016online, shou2018online, xu2021long,xu2019temporal,wang2021oadtr,zhao2022progressive,eun2020learning,chen2022gatehub,cao2023e2e,an2023miniroad}, and action localization~\cite{kang2021cag,reza2025hat,kim2022sliding,yang2024active}.
To our knowledge, the only previous online video temporal grounding (OnVTG) method was proposed by \citet{gan2023temporal}. \citet{gan2023temporal} stores historical frames to predict the probability of the current frame being the start or end timestamps of the target event and uses a teacher network that can observe future frames for knowledge distillation. However, the target event can have various durations, and such frame-level memory and per-frame score prediction methods lack effective event modeling. In this paper, we propose an event-based OnVTG framework to model complete event-level information and make predictions based on event proposals.

Online video action detection predicts an action label for each frame in real time. Online video action localization further requires the output of the action's start and end timestamps. TeSTra uses $M$ learnable queries to compress history information through cross-attention~\cite{vaswani2017attention}. MiniROAD directly uses GRU~\cite{chung2014empirical} to encode history information.
OAT~\cite{kim2022sliding} and HAT~\cite{reza2025hat} introduce proposal-based methods for online video action localization, utilizing a memory to retain historical frames, employing a sliding window to generate action proposals, and further classifying these proposals. 
However, they do not account for the delay in predicting the start time, which cannot be overlooked in the OnVTG task, where the target segments are longer.
Besides, different from action detection and localization, which independently predict each occurring action, the OnVTG requires consideration of inter-event relationships described in the query. This necessitates a robust capability to handle historical information over long periods.
Therefore, we propose a hierarchical event memory that retains historical events at different scales to obtain long-term historical information. We also propose a future prediction branch to address the delay in model predictions.

\subsection{Memory for Long Video Understanding}

Some works~\cite{he2024ma,zhang2024flash,song2024moviechat,weng2024longvlm,wang2024videollamb,cheng2024enhancing,qian2024streaming} have explored how to enable video large language models to support long video inputs through memory mechanisms.
For example, MA-LMM~\cite{he2024ma} and MovieChat~\cite{song2024moviechat} store historical frame features using memory and compress them by merging similar frames. 
VideoLLaMB~\cite{wang2024videollamb} and VideoStreaming~\cite{qian2024streaming} use attention mechanisms to compress historical information into a fixed number of historical tokens. Flash-Vstream~\cite{zhang2024flash} and HEM-LLM~\cite{cheng2024enhancing} combine multiple methods to construct multiple memories. 
However, the memory used in these methods is not optimal for online video temporal grounding. On one hand, OnVTG requires a multi-granularity temporal understanding ability, while its memory only forms sequential time representations and fails to capture hierarchical event structures. For instance, they cannot represent sub-events within a larger event. In contrast, we propose a hierarchical event memory that can preserve different granularities of the same event. 
On the other hand, these methods tend to compress all the video frames into memory, as a user can ask about everything everywhere. However, in the online setting, they cannot store all frames due to memory constraints, and thus progressively compress fine-grained visual details during continuous video processing. This compression-induced information loss is particularly detrimental for temporal localization tasks requiring precise moment retrieval. 
Our hierarchical memory effectively addresses these issues: the lower-level memory only retains recent details, while the higher-level memory preserves long-term historical information.

\begin{figure*}
    \centering
    \includegraphics[width=0.9\linewidth]{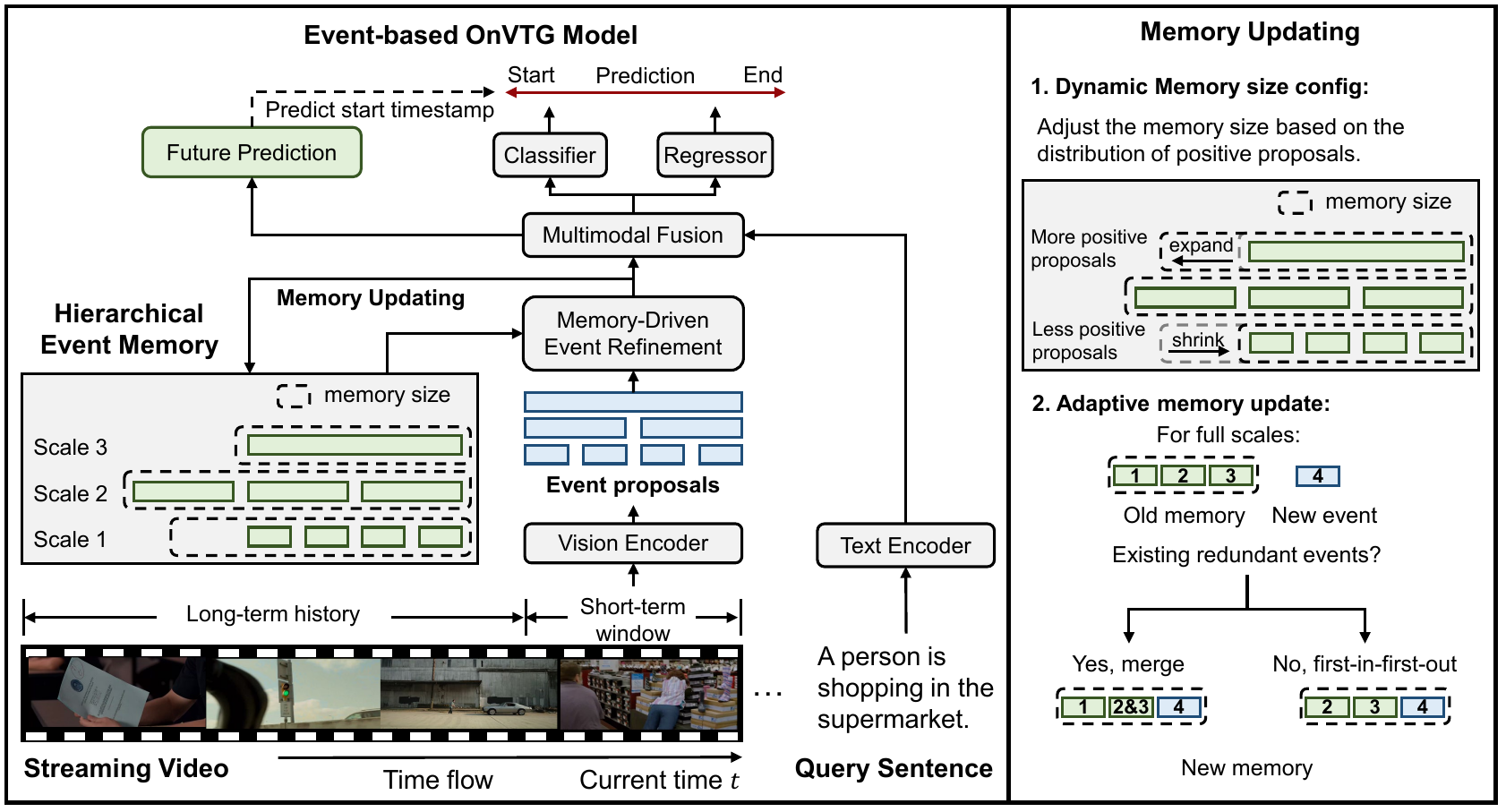}
    \caption{The framework of our proposed event-based online VTG method. \textbf{Left:} Our model processes a streaming video and a query sentence to locate a described event in real-time. It generates event proposals from a short-term window and enriches them with long-term context from the \textbf{Hierarchical Event Memory}, which stores historical events of varying durations. The model then outputs the event's boundaries: a \textbf{Future Prediction} branch provides a low-latency start prediction, while the refined proposal is used to determine the more accurate event boundaries. \textbf{Right:} The Hierarchical Event Memory is maintained through an efficient \textbf{Memory Updating} process, which uses a Dynamic Memory size config and Adaptive memory update rule to preserve the most relevant historical information.}
    
    \label{fig:method}
\end{figure*}

\begin{figure}
    \centering
    \includegraphics[width=0.9\linewidth]{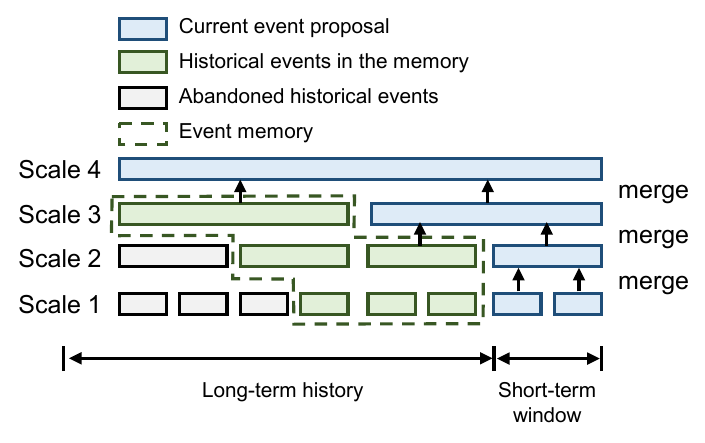}
    \caption{The construction of our hierarchical events. We use a segment tree structure to generate event proposals, where event proposals are divided into $L$ scales, and the large-scale proposals are obtained by merging the two adjacent small-scale proposals.}
    \label{fig:proposals}
\end{figure}

\begin{figure}
    \centering
    \includegraphics[width=\linewidth]{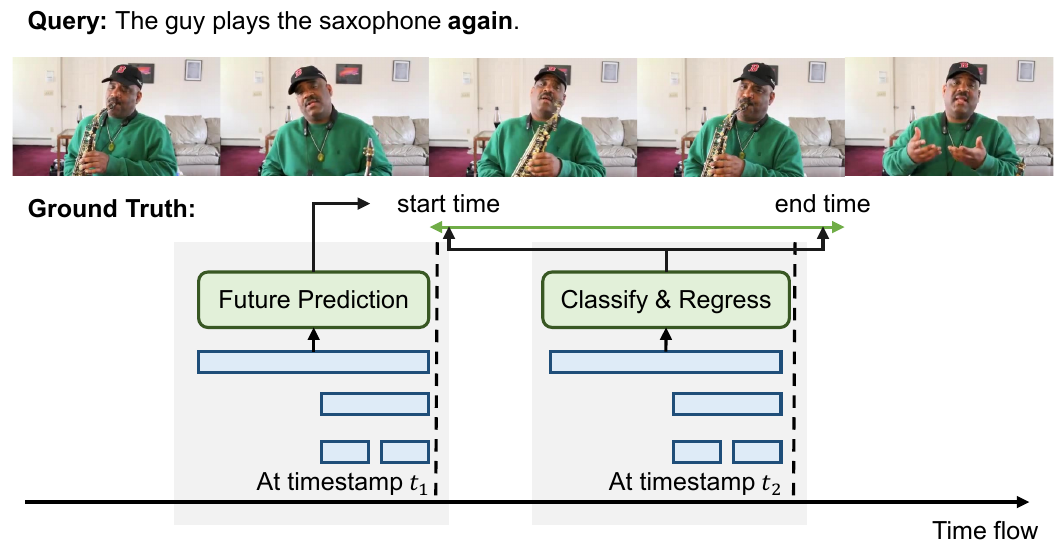}
    \caption{The illustration of future prediction. When the target event is about to start (at timestamp $t_1$), we can obtain real-time predictions of the start time. When the event is about to end (at timestamp $t_2$), we can obtain real-time predictions of the event's end time and provide a more accurate start time from event proposals.}
    \label{fig:future}
\end{figure}

\section{Method}
\label{sec:method}

\subsection{Problem Definition}

Online video temporal grounding (OnVTG) aims to localize the target event $(s,e)$ related to a given natural language query $Q$ in an untrimmed video $V=\{v_i\}_{i=1}^T$, where $s$ and $e$ are the start and end time of the target event and $T$ is the number of frames in the video. The online constraint of OnVTG requires that at any timestamp $t (1\le t\le T)$, only partial video $V=\{v_i\}_{i=1}^t$ is observed by the model, and the model should predict the target event as early as possible.

\subsection{Overview}
The pipeline of our method is illustrated in Figure~\ref{fig:method}. To ensure real-time predictions, we construct event proposals that end within the current short-term window and promptly predict which proposals match the query text. Since target events can have various durations and may require long-term historical information for accurate localization, we propose a hierarchical event memory to store historical events of varying durations and distances in time to refine the current event proposals.

Specifically, as illustrated in Figure~\ref{fig:proposals}, we use a segment tree structure to generate event proposals with various durations (colored in blue) based on the current short-term window and the event memory. The event proposals are divided into $L$ scales, and the large-scale proposals are obtained by merging the two adjacent small-scale proposals. To provide long-term historical information, we further use history memory to refine the current event proposals. As online videos are streaming inputs and can go on indefinitely, we need to retain more valuable and low-redundancy historical information within a limited memory size.
The scale that aligns with the typical duration of events is more likely to preserve valuable event information, thus, we propose a dynamic memory size configuration for different scales, ensuring that those scales with more positive events have larger memory sizes. We also propose an adaptive memory update rule to merge redundant events.

To enable a real-time prediction of the start time, we propose a future prediction branch in Figure~\ref{fig:future}, which requires each event proposal to predict whether the target event will start soon. When the target event is about to start (at $t_1$ in Figure~\ref{fig:future}), we can obtain real-time predictions of the start time. When the event is about to end (at $t_2$ in Figure~\ref{fig:future}), we can obtain real-time predictions of the event's end time and provide a more accurate start time from event proposals.

\subsection{Event-based OnVTG Model}
\label{sec:model}

In video temporal grounding, target events can have various durations and may require long-term historical information for accurate localization. In this paper, we propose a novel hierarchical event based OnVTG pipeline that utilizes the segment tree structure to generate event proposals with various durations and make predictions based on these event proposals.

\textbf{Hierarchical Event Construction.} As shown by the blue proposals in Figures \ref{fig:method} and \ref{fig:proposals}, to ensure the real-time prediction, we construct event proposals that end within the short window \([t - L_s, t]\) at time \( t \), where $L_s$ is the short-window size. The recent historical events are also stored in an event memory $M$ (colored in green) to preserve long-term historical information. As illustrated in Figure~\ref{fig:proposals}, we use a segment tree structure to generate event proposals to reduce event redundancy and improve efficiency.  
Specifically, for the short window frames $V=\{v_i\}_{i=t-L_s+1}^{t}$, we use a pre-trained vision encoder~\cite{vaswani2017attention} to extract the visual features and use 1-D convolution to obtain the proposal features at the first scale: $P^1=\{p^1_{i}\}_{i=t-L_s+1}^{t}$. Then, for the $(j+1)$-th scale, proposals are obtained by merging two adjacent proposals at the $j$-th scale $(1 \le j \le L-1)$:
\begin{equation}
\begin{aligned}
    p^{j+1}_{i} = \text{MLP}([p^{j}_{2i};p^{j}_{2i-1}]) \\
    \label{eq: proposal}
\end{aligned}
\end{equation}
where $[;]$ represents the concatenation and $\text{MLP}(\cdot)$ is the multilayer perceptron~\cite{haykin1994neural}. At the $j$-th scale, the duration of proposal $P^{j+1}$ is $2^j$. When $j$ is large, this duration may exceed the length of the short-term window (for example, scales 3 and 4 in Figure~\ref{fig:proposals}). As shown in Figure~\ref{fig:proposals}, in this case, the most recent historical event stored in memory at scale $j$ will be involved in the calculation of Eq(\ref{eq: proposal}).

\textbf{Memory-Driven Event Refinement.} For some queries like `the guy plays the saxophone again', the model requires other events information that may happen early to assist in the localization of the current query. Thus, we further refine the current event proposals by the historical events stored in the memory $M$: $P^j = \text{E}(P^j; M)$, where $\text{E}(\cdot)$ is the transformer encoder layer~\cite{vaswani2017attention}.
The small-scale memory provides fine-grained recent information, while the large-scale memory retains long-term coarse-grained information.

\textbf{Proposal-based Model Prediction.} After obtaining the event proposal features $P$, we encode the query sentence $Q$ using a text encoder and fuse the event proposal features with the query features with transformer decoder: $\hat{P} = \text{D}(P, \text{E}_{text}(Q))$. Then, for each event proposal $\hat{p}_i$, we use a classifier to determine whether it is a positive event: $c_i = \text{MLP}(\hat{p}_i)$. An event proposal is regarded as positive if the IoU between the proposal and ground truth is larger than a certain threshold. For the positive proposal, we further regress the offsets between the start/end times of the GT and the proposal to obtain more accurate predictions: $o^s_i, o^e_i = \text{MLP}(\hat{p}_i)$. We use focal loss~\cite{lin2017focal} to supervise the classification results and use DIoU loss~\cite{zheng10faster} to supervise the regression results:
\begin{equation}
    \mathcal{L}_{cls} = \frac{1}{N}\sum_{i=1}^{N}\mathcal{L}_{focal}(c_i, \hat{c}_i)
\end{equation}
\begin{equation}
    \mathcal{L}_{reg} = \frac{1}{N}\sum_{i=1}^{N}\mathcal{L}_{DIoU}(p^s_i+o^s_i, p^e_i+o^e_i, s, e) \mathbf{1}_{\{\hat{c}_i = 1\}}
\end{equation}
where $\hat{c}_i=1$ if and only if event proposal $\hat{p}_i$ is positive, $(p^s_i, p^e_i)$ is the start and end time of proposal $\hat{p}_i$, $(s, e)$ is the start and end time of the ground truth, $\mathbf{1}_{\{\cdot\}}$ is the indicator function. 

\textbf{Future Prediction.}
Relying solely on event proposals for predictions can lead to delays in predicting the start of events since the model can only obtain a complete event proposal when the event is about to end. To address this issue, we propose the future prediction branch. Specifically, at time $t$, we use an MLP to predict the probability $c^f_t$ that the target event starts in a future window $[t+a, t+b]$ using the current event proposals $\hat{P}$, where $(a,b)$ are hyperparameters. Then, we further predict the offset $o^f_t$ by another MLP. The future prediction loss is:
\begin{equation}
    \mathcal{L}_{future} = \mathcal{L}_{focal}(c^f_t, \hat{c}^f_t) + \|o^f_t-(s -t)\|_1 \mathbf{1}_{\{\hat{c}^f_t = 1\}}
\end{equation}
where $s$ is the start time of the ground truth, $\hat{c}^f_t=1$ if and only if $t+a\le s \le t+b$.
As illustrated in Figure~\ref{fig:future}, when the target event is about to start (at $t_1$), we can obtain real-time predictions of the start time. When the event is about to end (at $t_2$), we can obtain real-time predictions of the event's end time and provide a more accurate start time from event proposals.

\subsection{Memory Updating}
\label{sec:memory}

 When the next short-term window comes, the event proposals in the current window will be added to the memory. As online videos are streaming inputs and can go on indefinitely, we need to retain more valuable and low-redundancy historical information within a limited memory size. Thus, we propose a dynamic memory size configuration and an adaptive memory update rule.

\textbf{Dynamic memory size configuration.}
The scale that aligns with the typical duration of events is more likely to preserve valuable event information.
Thus, we only specify the total size of the memory as $K$, while the size of the memory at each scale depends on the probability of positive sample events occurring at that scale. Specifically, for each scale, we count $w_i$ to represent the frequency of positive samples at that scale.
We ensure that the memory size for each scale is at least 1 to successfully retrieve historical events when constructing the current event proposal in Figure~\ref{fig:proposals}. The remaining memory size will be allocated proportionally according to $w_i$:
\begin{equation}
    K_i = 1 + (K - L)\frac{w_i}{\sum_{j=1}^Lw_j}, i=1,2,...,L
\end{equation}
where $L$ is the number of scales.

\textbf{Adaptive memory update.}
When a new event proposal is added to memory, in some scale, the memory may exceed the size $K_i$. Previous methods typically employed a first-in, first-out update rule. However, when long-duration similar segments appear in the video, memory can easily become filled with redundant content. Therefore, we propose an adaptive memory update rule: For the scale $i$ which exceeds the memory size, we calculate the cosine similarity between two adjacent events in scale $i$. If there are adjacent events with a similarity greater than $\delta$, we merge them through average pooling, repeating this process until the memory size does not exceed $K_i$. If there are no adjacent events with a similarity greater than $\delta$, we follow a first-in-first-out principle, retaining the most recent $K_i$ events in the memory.

\subsection{Training and inference}

\textbf{Training.} For training efficiency, we group queries from the same video into the same batch. After obtaining event proposals for all the short-term windows in the video, we fuse them with text for prediction and calculate classification loss $\mathcal{L}_{cls}$, regression loss $\mathcal{L}_{reg}$, and future prediction loss $\mathcal{L}_{future}$. The total loss is:
\begin{equation}
    \mathcal{L} = \mathcal{L}_{cls} + \mathcal{L}_{reg} + \mathcal{L}_{future}
\end{equation}

\noindent \textbf{Inference.} Our future prediction branch enables us to infer in two ways: one is to predict the start and end times based on the event proposal, and the other is to predict the start time using the future prediction branch and the end time using the event proposal. The former achieves higher performance by observing the entire event before making predictions, while the latter can make predictions when the event begins, achieving less start time prediction latency. In Section~\ref{sec:sota} of our experiments, we report the performance under both inference methods.

\subsection{Discussion}
Here we discuss the advantages of our hierarchical event memory over previous frame-level memory in HAT~\cite{reza2025hat}, OAT~\cite{kim2022sliding}, and \citet{gan2023temporal}. When the memory size is fixed to $K$, frame-level memory can only retain the recent $K$ frames of historical information. In the current window, the model can only construct event proposals with a maximum duration of $K$, making it difficult for frame-level memory to locate such long-term events.
Our hierarchical event memory, by utilizing a segment tree structure to construct events, as shown in Figure~\ref{fig:proposals}, allows the model to obtain event proposals of length $2^K$ (a total of $K$ scales, with each scale's memory storing one latest event). In addition, our small-scale memory can retain fine-grained but recent information, while the large-scale memory retains long-term but coarse-grained information to meet the needs of locating target events of different durations.
\section{Experiment}
\label{sec:experiment}

\subsection{Datasets}
\label{sec:datasets}

We follow the settings of \citet{gan2023temporal} and conduct experiments on three datasets, MAD~\cite{soldan2022mad} and ActivityNet Captions~\cite{krishna2017dense} and TACoS~\cite{rohrbach2014coherent}.
\textbf{MAD}~\cite{soldan2022mad} is a large dataset with long videos.  It comprises over 1,200 hours of video and 384,000 queries from high-quality mainstream movie audio descriptions. The average duration of videos is 110.8 minutes, while the average duration of the target segments is only 4.1 seconds, making this dataset challenging.
\textbf{ActivityNet Captions}~\cite{krishna2017dense} comprises 14,926 videos and 71,953 natural language queries about human activity.  The average duration of videos is 117.6 seconds, and the average duration of the target segments is 37.1 seconds.
\textbf{TACoS}~\cite{rohrbach2014coherent} comprises 127 videos and 18,227 natural language queries. The average duration of videos is 286.59 seconds, and the average duration of the target segments is 27.88 seconds.

\subsection{Metrics}

We follow \citet{gan2023temporal} to use an offline evaluation (i.e. evaluate after all videos have been input) and use the metric “R@$n$,IoU=$m$ ($R^n_m$)” to evaluate the model. $R^n_m$ calculates the percentage of the model's top-$n$ predictions that have at least one prediction having an IoU with the ground truth greater than $m$. 
Inspired by OAT~\cite{kim2022sliding}, we also evaluate the model's start time prediction delay (SD) and end time prediction delay (ED). The SD/ED indicates the difference between the timestamp when the model makes a prediction and the ground-truth start/end time. Smaller values are better, and negative values mean the model can predict the event before it actually occurs. Details are in the Supplementary Materials.


\subsection{Implementation Details}
We follow previous works\cite{gan2023temporal} to use C3D~\cite{tran2015learning} to extract frame features on ActivityNet Captions and TACoS datasets and use CLIP~\cite{radford2021clip} to frame features on the MAD dataset. The hyperparameters are $a=-4, b=4, L=8$, the length of the short-term window is $8$, and the total size of the memory is $64$ on all datasets. We trained the model with the AdamW~\cite{loshchilov2017adamw} optimizer with a learning rate of $0.002$ on ActivityNet Caption, $0.0001$ on MAD, and $0.001$ on TACoS. 

\begin{table}[t]
    \centering
    \scalebox{0.8}{
    \begin{tabular}{c|cccc|cc}
    \toprule
         Method& $R^1_{0.5}$ & $R^1_{0.7}$&$R^5_{0.5}$& $R^5_{0.7}$ & SD & ED \\
    \midrule
        \multicolumn{7}{c}{\textcolor{gray}{\textbf{Offline video temporal grounding} methods}}\\
    \midrule
        \textcolor{gray}{G2L}~\cite{li2023g2l} &  \textcolor{gray}{42.74} &\textcolor{gray}{30.95} &\textcolor{gray}{65.83} &\textcolor{gray}{49.86} &\textcolor{gray}{-} & \textcolor{gray}{-}\\
        \textcolor{gray}{SnAG}~\cite{mu2024snag} &  \textcolor{gray}{56.44 } &\textcolor{gray}{44.86 } &\textcolor{gray}{ 81.15} &\textcolor{gray}{70.66} &\textcolor{gray}{-} & \textcolor{gray}{-}\\
    
    \midrule
        \multicolumn{7}{c}{\textbf{Online action detection} methods~\hyperlink{foot1}{\textsuperscript{1}}}\\
    \midrule
        OadTR\cite{wang2021oadtr}& 21.12& 10.92& 37.99 &21.09& - &-\\
        LSTR\cite{xu2021long}&  26.02& 16.75& 43.01& 27.99 &- &-\\
        GateHUB\cite{chen2022gatehub} &27.10 &17.25& 43.44 &26.87 &- & -\\
        TeSTra\cite{zhao2022testra} &27.43 &16.84& 43.57 &27.11 &0.89s & 1.14s\\
        MiniROAD\cite{an2023miniroad} &28.03 &17.98& 44.11 &27.64 &1.56s & 1.08s\\
    \midrule
        \multicolumn{7}{c}{\textbf{Online action localization} methods~\hyperlink{foot1}{\textsuperscript{1}}}\\
    \midrule
        OAT\cite{kim2022sliding} & 32.53 & 12.11 & 50.96& 33.14 & 18.11s & -1.09s\\
        HAT\cite{reza2025hat} & 34.15 & 14.53 & 51.16& 34.98 & 20.07s & \textbf{-1.98s}\\
    \midrule
        \multicolumn{7}{c}{\textbf{Online video temporal grounding} methods}\\
    \midrule
        \citet{gan2023temporal} & 29.74 & 19.07& 48.11 &31.19& 1.42s & 1.38s \\
        \textbf{Ours w/ FP} &37.44	&27.32	&57.49	&44.44	&\textbf{-1.28s}	&-3.78s \\
        \textbf{Ours w/o FP} & \textbf{44.19} &\textbf{30.87}	&\textbf{68.96}	&\textbf{52.69}	&22.38s	&\textbf{-4.26s} \\
    \bottomrule
    \end{tabular}}
    \caption{Performance comparison on TACoS dataset. `FP' represents the future prediction branch.}
    \label{tab:tacos}
\end{table}

\begin{table}[t]
    \centering
    \tabcolsep=1.9mm
    \scalebox{0.8}{
    \begin{tabular}{c|cccc|cc}
    \toprule
         Method& $R^1_{0.5}$ & $R^1_{0.7}$&$R^5_{0.5}$& $R^5_{0.7}$ & SD & ED \\
    \midrule
        \multicolumn{7}{c}{\textcolor{gray}{\textbf{Offline video temporal grounding} methods}}\\
    \midrule
        \textcolor{gray}{G2L}~\cite{li2023g2l} &  \textcolor{gray}{51.68 } &\textcolor{gray}{33.35} &\textcolor{gray}{ 81.32 } &\textcolor{gray}{67.60} &\textcolor{gray}{-} & \textcolor{gray}{-}\\
        \textcolor{gray}{SnAG}~\cite{mu2024snag} &  \textcolor{gray}{48.55 } &\textcolor{gray}{30.56  } &\textcolor{gray}{81.71} &\textcolor{gray}{ 63.41 } &\textcolor{gray}{-} & \textcolor{gray}{-}\\
    \midrule
        \multicolumn{7}{c}{\textbf{Online action detection} methods~\hyperlink{foot1}{\textsuperscript{1}}}\\
    \midrule
        OadTR\cite{wang2021oadtr}& 23.27 &10.97 &48.15 &29.76& - &-\\
        LSTR\cite{xu2021long}& 24.05 &11.19& 50.77 &31.63 &- &-\\
        GateHUB\cite{chen2022gatehub} &23.30 &11.31& 50.25& 31.00 &- &- \\
        TeSTra\cite{zhao2022testra} &24.14 &11.69& 52.83 &32.47 &1.86s & 1.14s\\
        MiniROAD\cite{an2023miniroad} &24.31 &12.09& 53.10 &33.13 &1.09s & 1.23s\\
    \midrule
        \multicolumn{7}{c}{\textbf{Online action localization} methods~\hyperlink{foot1}{\textsuperscript{1}}}\\
    \midrule
        OAT\cite{kim2022sliding} & 34.41 & 14.37 & 53.41& 38.49 & 30.07s & -6.41s\\
        HAT\cite{reza2025hat} & 36.56 & 16.64 & 55.69& 40.51 & 31.41s & -7.41s\\
    \midrule
        \multicolumn{7}{c}{\textbf{Online video temporal grounding} methods}\\
    \midrule
        \citet{gan2023temporal} & 25.48 &12.56 &53.77 &33.70& 2.13s & 1.49s \\
        \textbf{\makecell{Ours w FP}} & 42.89	&24.49	&67.82	&51.74	&\textbf{-1.58s}	&\textbf{-10.96s} \\
        \textbf{\makecell{Ours w/o FP}} & \textbf{45.29}&	\textbf{26.25}	&\textbf{76.24	}&\textbf{62.14}	&41.10s	&-10.89s \\
    \bottomrule
    \end{tabular}}
    \caption{Performance comparison on ActivityNet Captions dataset. `FP' represents the future prediction branch.}
    \label{tab:anet}
\end{table}

\begin{table}[t]
    \centering
    \scalebox{0.8}{
    \begin{tabular}{c|cccc|cc}
    \toprule
         Method& $R^5_{0.3}$ & $R^5_{0.5}$&$R^{50}_{0.3}$& $R^{50}_{0.5}$ & SD & ED \\
    \midrule
        \multicolumn{7}{c}{\textcolor{gray}{\textbf{Offline video temporal grounding} methods}}\\
    \midrule
        \textcolor{gray}{SOONet}~\cite{pan2023scanning} &  \textcolor{gray}{19.64  } &\textcolor{gray}{13.14} &\textcolor{gray}{ 44.78 } &\textcolor{gray}{ 32.59} &\textcolor{gray}{-} & \textcolor{gray}{-}\\
        \textcolor{gray}{SnAG}~\cite{mu2024snag} &  \textcolor{gray}{ 20.60  } &\textcolor{gray}{13.75  } &\textcolor{gray}{46.68 } &\textcolor{gray}{ 35.24 } &\textcolor{gray}{-} & \textcolor{gray}{-}\\
    \midrule
        \multicolumn{7}{c}{\textbf{Online action detection} methods~\hyperlink{foot1}{\textsuperscript{1}}}\\
    \midrule
        OadTR\cite{wang2021oadtr}&  2.50 &0.90 &8.61 &4.12& - &-\\
        LSTR\cite{xu2021long}&  3.56 &1.43 &11.73 &4.99 & -&-\\
        GateHUB\cite{chen2022gatehub} &3.38 &1.47& 11.96& 4.74 & -& -\\
        TeSTra\cite{zhao2022testra} &3.46 &1.33& 12.98 &6.61 &1.24s & 0.97s\\
        MiniROAD\cite{an2023miniroad} &4.58 &1.64& 14.07 &7.03 &1.51s & 1.13s\\
    \midrule
        \multicolumn{7}{c}{\textbf{Online action localization} methods~\hyperlink{foot1}{\textsuperscript{1}}}\\
    \midrule
        OAT & 6.17 & 4.41 & 20.11& 13.14 & 3.02s & -1.17s\\
        HAT & 7.14 & 5.11 & 22.03& 14.47 & 3.14s & -1.23s\\
    \midrule
        \multicolumn{7}{c}{\textbf{Online video temporal grounding} methods}\\
    \midrule
        \citet{gan2023temporal} & 4.71 &2.00 &16.34 &7.80& 0.13s & 1.52s \\
        \textbf{Ours w/ FP} & 9.84	&6.43	&16.67	&12.18	&\textbf{0.64s}	&-1.10s \\
        \textbf{Ours w/o FP} & \textbf{15.76}&	\textbf{11.07}&	\textbf{37.84}&	\textbf{29.21}&	3.60s&	\textbf{-1.45s} \\
    \bottomrule
    \end{tabular}}
    \caption{Performance comparison on MAD dataset. `FP' represents the future prediction branch.}
    \label{tab:mad}
\end{table}

\subsection{Comparison with Other Methods}
\label{sec:sota}

In Table~\ref{tab:tacos}~\ref{tab:anet}~\ref{tab:mad}, we compare the performance and prediction delay on the TACoS, ActivityNet Captions, and MAD datasets, respectively. We compare our method with the \textbf{online video temporal grounding} baseline proposed by \citet{gan2023temporal}. To our knowledge, this is the only work designed for online video temporal grounding. We also compare with \textbf{online action detection} and \textbf{online action localization} baselines. These methods are not originally designed for online video temporal grounding, and we have modified them for this task. 
For the online action detection baselines, we compare ours with OAT and HAT, both of which use the proposal-based framework. We introduce a text encoder and utilize cross-attention to fuse proposal features with text features, adapting them to video temporal grounding.

\textbf{Performance comparisons.} (1) In all datasets, the proposal-based methods (OAT, HAT, and ours) show significant advantages. For example, in Table~\ref{tab:tacos}, ours w/o FP outperforms \citet{gan2023temporal} by 14.45\% on $R^1_{0.5}$. (2) Among those proposal-based methods, ours w/o FP outperforms HAT by 10.04\% on $R^1_{0.5}$ in Table~\ref{tab:tacos}. 
This is because our hierarchical memory can retain longer historical information compared to the frame-level memory in HAT, allowing for the construction of proposals of varying durations. 

\textbf{Delay comparisons.} (1) The proposal-based methods (OAT, HAT, and ours) can predict the end time even before the event has concluded. This may be because the model has learned the priors of how long various events typically last. (2) Relying solely on proposals for prediction results in a delay in predicting the start time. For example, in Table~\ref{tab:tacos}, these methods have a delay of about 20 seconds when predicting the start time on TACoS. This is because the model only obtains a complete event proposal when the event is about to end.
(3) Our proposed future prediction branch (ours w/ FP) can reduce the delay, making predictions even before the event is about to start. However, the accuracy decreases when applying future prediction (on TACoS, $R^1_{0.5}$ drops by 6.75\% compared to ours w/o FP), but it is still higher than the existing baseline. Whether to use future prediction depends on the application's demand for latency.

\hypertarget{foot1}{\footnotetext[1]{These methods are not originally designed for online video temporal grounding, and we have modified them for this task.}}

\begin{table}[t]
    \centering
    \tabcolsep=1mm
    \scalebox{0.8}{
    \begin{tabular}{cccc|cccc}
    \toprule
         \makecell{Event \\memory}& \makecell{Dynamic \\size}& \makecell{Adaptive \\updating}& \makecell{Future \\prediction} & $R^1_{0.5}$ & $R^1_{0.7}$ & SD & ED \\
    \midrule
        \XSolidBrush & \XSolidBrush & \XSolidBrush & \XSolidBrush & 34.88 & 23.47 & 18.36s &-3.17s \\
        \CheckmarkBold &  & &  & 41.61 & 29.84 & 20.59s &-4.18s \\
        \CheckmarkBold & \CheckmarkBold & & & 43.97 & 30.54 & 22.41s &-4.16s \\
        \CheckmarkBold & \CheckmarkBold & \CheckmarkBold & & \textbf{44.19} & \textbf{30.87} & 22.38s &\textbf{-4.26s} \\
        \CheckmarkBold & \CheckmarkBold & \CheckmarkBold & \CheckmarkBold &37.44 &27.32 & \textbf{-1.28s} & -3.78s \\
    \bottomrule
    \end{tabular}}
    \caption{Ablations of each component on TACoS dataset.}
    \label{tab:ab_comp}
\end{table}

\begin{figure}
    \centering
    \includegraphics[width=0.6\linewidth]{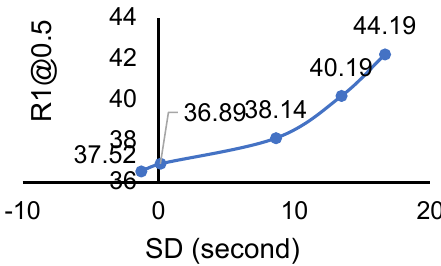}
    \caption{Accuracy-latency trade-off on TACoS dataset.}
    \label{fig:trade-off}
\end{figure}


\begin{table}[t]
    \centering
    \scalebox{1.0}{
    \begin{tabular}{c|cc|c}
    \toprule
         Method & param.  & Inference speed&$R^5_{0.3}$ \\
    \midrule
        \citet{gan2023temporal} & 29.5M& 734.3 FPS &4.71\\
        Ours & 36.2M & 614.5 FPS &15.76\\
    \bottomrule
    \end{tabular}}
    \caption{Number of parameters and inference speed comparison on MAD dataset.}
    \label{tab:ab_speed}
\end{table}

\subsection{Ablation Studies}

We perform ablation studies on the TACoS dataset to analyze the effectiveness of the proposed method.

\textbf{Effectiveness of each component.} In Table~\ref{tab:ab_comp}, we analyze the effectiveness of different modules. For the model without event memory, we used a frame-level memory to retain the latest $K$ frame features, where $K$ is the size of the memory. For the model without dynamic size, we equally set the memory size across each scale. For the model without adaptive updating, we used a first-in-first-out strategy to update the memory. Our event memory, dynamic size, and adaptive updating can enhance the model's performance, demonstrating the effectiveness of these modules. Our future prediction can significantly reduce the prediction delay at the start time, but it also leads to a decrease in performance. Whether to use future prediction depends on the application's demand for latency.


\textbf{Accuracy-latency trade off.} Our method offers a flexible trade-off by combining future and event-based predictions. For early start-time estimation, future prediction is used, while event-based prediction refines the start-time and determines the end-time as the event concludes.
In addition, the performance and latency can be adjusted via the future prediction window  $(a, b)$ introduced in our future prediction branch. When $a < 0$, the model is allowed to make predictions shortly after an event starts, thus improving the performance while increasing the prediction delay. In \cref{fig:trade-off}, we provide the accuracy and delay when adjusting the future prediction window, which helps to select models based on latency tolerance.


\textbf{Speed Comparison.} In Table~\ref{tab:ab_speed}, we compare the speed and the number of parameters of our model with the baseline~\cite{gan2023temporal}. This excludes the visual and text feature extraction, as we use the same pre-extracted features. Our method has similar parameters to the baseline. In terms of inference speed, both models meet the real-time requirements for online video (exceeding the frame rate of typical videos). Our method is 17\% slower than the baseline, but the performance of $R^5_{0.3}$ is 3.3 times that of the baseline.


\begin{figure}
    \centering
    \includegraphics[width=\linewidth]{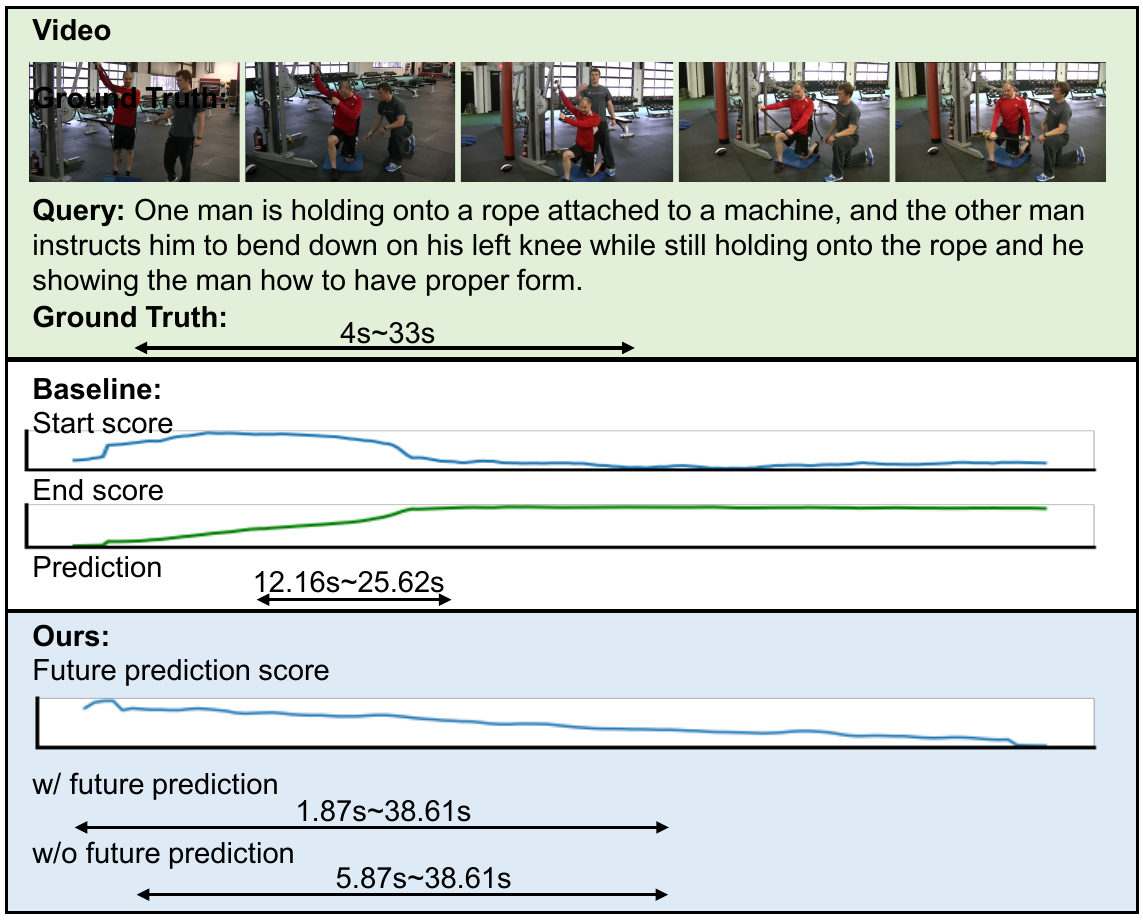}
    \caption{Qualitative results on the ActivityNet Captions dataset.}
    \label{fig:vis}
\end{figure}

\textbf{Qualitative Comparison.} Figure~\ref{fig:vis} provides visualizations between our method and \citet{gan2023temporal}. The baseline utilizes frame-level historical information, but when the target event has a long duration, the model's predictions are not accurate. 
In contrast, our method, by storing hierarchical historical events, is able to make more accurate predictions. 
Our future prediction branch can also predict the start time of future events. The prediction is more accurate from the event proposal when the event is about to end.

\section{Conclusion}
\label{sec:conclusion}

In this work, we propose a hierarchical event memory for online video temporal grounding. 
We introduce a hierarchical event memory that retains historical events of varying durations to preserve historically valuable event information. We use adaptive memory updating to reduce the redundant proposal in the memory and propose the dynamic configuration method of memory sizes to enhance the utilization efficiency of the memory. To address the issue of delays in model predictions, we propose a future prediction branch. Experiments on MAD, ActivityNet, and TACoS datasets demonstrate the effectiveness of our method.


\noindent\textbf{Acknowledgements.} This work was supported by the
grants from the National Natural Science Foundation of
China (62372014, 62525201, 62132001, 62432001), Beijing Nova Program and Beijing Natural Science Foundation (4252040, L247006).

{
    \small
    \bibliographystyle{ieeenat_fullname}
    \bibliography{main}
}

\end{document}